\documentclass{article}

\usepackage{arxiv}

\usepackage[utf8]{inputenc} 
\usepackage[T1]{fontenc}    
\usepackage{hyperref}       
\usepackage{url}            
\usepackage{booktabs}       
\usepackage{amsfonts}       
\usepackage{nicefrac}       
\usepackage{microtype}      
\usepackage{lipsum}
\usepackage{graphicx}
\usepackage{graphics}
\graphicspath{ {./images/} }

\usepackage{amsmath} 
\usepackage{amssymb}  
\usepackage{booktabs}
\usepackage{siunitx}
\usepackage{multirow}
\usepackage[dvipsnames]{xcolor}
\usepackage{hyperref}
\usepackage{tabularx}

\usepackage[export]{adjustbox}
\usepackage{algorithm}
\usepackage{algpseudocode}
\usepackage{marvosym}

\title{GPS-free Autonomous Navigation in Cluttered Tree Rows with Deep Semantic Segmentation}

\author{Alessandro~Navone$^{1}$, Mauro~Martini$^{1}$, Marco~Ambrosio$ {1}$, Andrea~Ostuni${1}$, Simone Angarano$^{1}$ and Marcello Chiaberge$^{1}$ 
\thanks{$^{1}$ Department of Electronics and Telecommunications, Politecnico di Torino, 10129, Torino, Italy. \tt\footnotesize \{firstname.lastname\}@polito.it}}

\author{
 Alessandro Navone \\
  Department of Electronics and Telecommunications \\
  Politecnico di Torino\\
  Torino, TO, 10129 \\
  \texttt{alessandro.navone@polito.it} \\
   \And
 Mauro Martini \\
  Department of Electronics and Telecommunications \\
  Politecnico di Torino\\
  Torino, TO, 10129 \\
  \texttt{mauro.martini@polito.it} \\
   \And
Marco Ambrosio\\
  Department of Electronics and Telecommunications \\
  Politecnico di Torino\\
  Torino, TO, 10129 \\
  \texttt{marco.ambrosio@polito.it} \\
   \And
 Andrea Ostuni \\
  Department of Electronics and Telecommunications \\
  Politecnico di Torino\\
  Torino, TO, 10129 \\
  \texttt{andrea.ostuni@polito.it} \\
  \And
 Simone Angarano \\
  Department of Electronics and Telecommunications \\
  Politecnico di Torino\\
  Torino, TO, 10129 \\
  \texttt{simone.angarano@polito.it} \\
  \And
 Marcello Chiaberge \\
  Department of Electronics and Telecommunications \\
  Politecnico di Torino\\
  Torino, TO, 10129 \\
  \texttt{marcello.chiaberge@polito.it} \\
}

\begin{document}
\maketitle
\begin{abstract}
Segmentation-based autonomous navigation has recently been presented as an appealing approach to guiding robotic platforms through crop rows without requiring perfect GPS localization. Nevertheless, current techniques are restricted to situations where the distinct separation between the plants and the sky allows for the identification of the row's center. However, tall, dense vegetation, such as high tree rows and orchards, is the primary cause of GPS signal blockage. In this study, we increase the overall robustness and adaptability of the control algorithm by extending the segmentation-based robotic guiding to those cases where canopies and branches occlude the sky and prevent the utilization of GPS and earlier approaches. An efficient Deep Neural Network architecture has been used to address semantic segmentation, performing the training with synthetic data only. Numerous vineyards and tree fields have undergone extensive testing in both simulation and real-world to show the solution's competitive benefits.
\end{abstract}


\section{Introduction}\label{sec:intro}
Precision agriculture has been pushing technological limits to maximize crop yield, boost agricultural operations efficiency, and minimize waste \cite{zhai2020decision}. Contemporary agricultural systems need to be able to gather synthetic essential information from the environment, make or recommend the best decisions based on that knowledge, and carry out those decisions with extreme speed and precision. Deep learning techniques have demonstrated considerable potential in creating these systems by evaluating data from numerous sources, enabling large-scale, high-resolution monitoring, and offering precise insights for both human and robotic actors. Recent developments in deep learning also offer competitive advantages for real-world applications, including generalization to unknown data \cite{martini2021domain, angarano2022back, angarano2023domain} and model optimization for quick inference on low-power embedded hardware \cite{mazzia2020real, angarano2021ultra}. Simultaneously, advancements in service robotics have made it possible for self-governing mobile agents to assume the role of artificial intelligence perception systems and collaborate with them to complete intricate tasks in unstructured settings \cite{holland2021service, eirale2022marvin}.

\begin{figure}[H]
    \setlength{\fboxsep}{0pt}
    \centering
    \resizebox{0.9\columnwidth}{!}{%
    \begin{tabular}{c c}
         \framebox{\includegraphics[width=0.332\columnwidth]{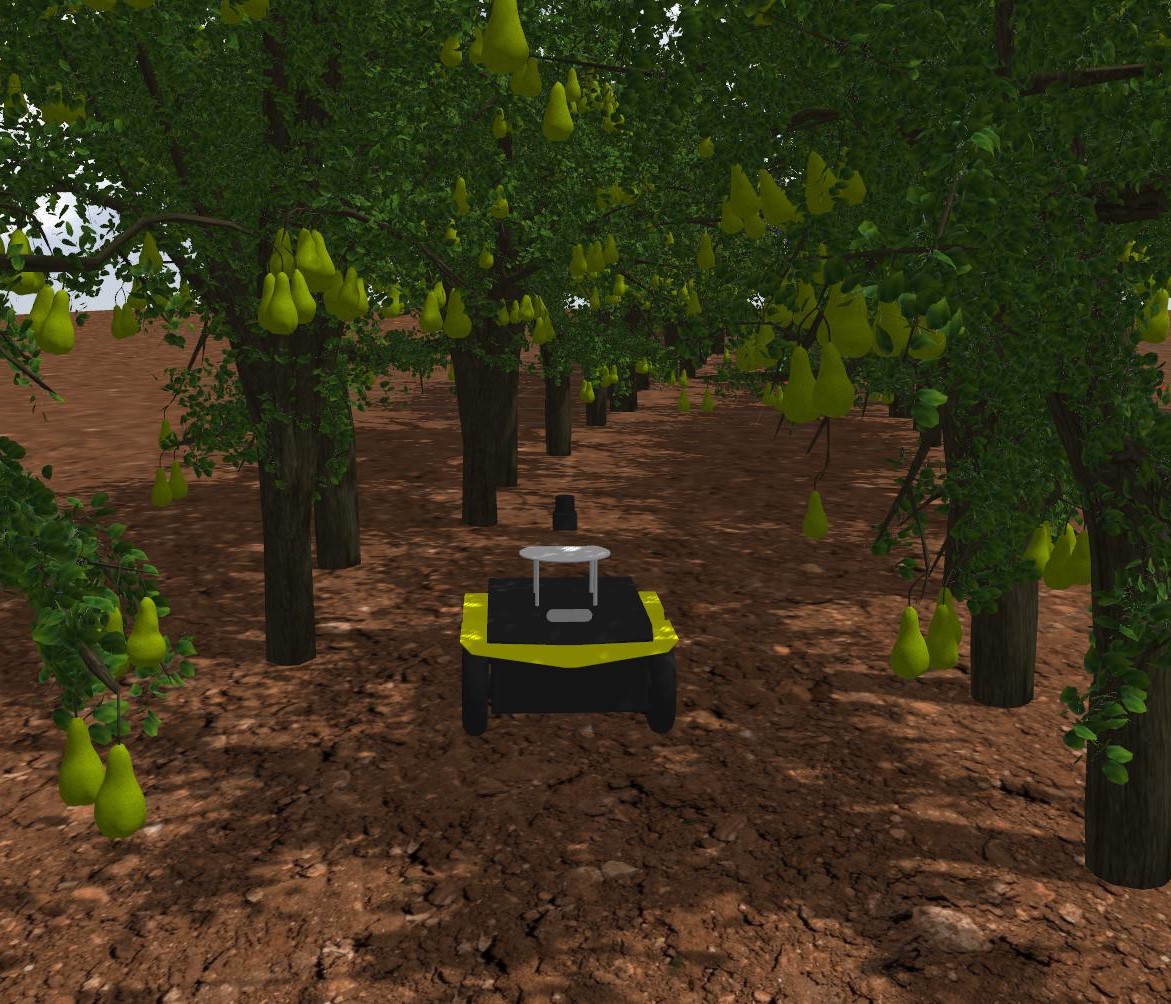}} & \framebox{\includegraphics[width=0.40\columnwidth]{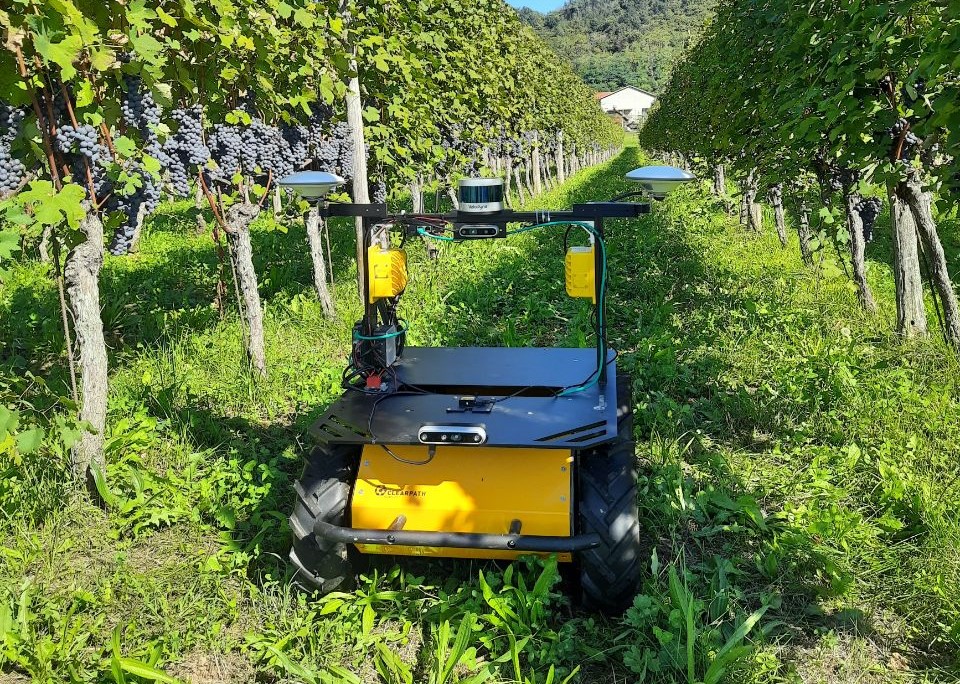}} 
    \end{tabular}
    }
    \caption{The proposed SegMin and SegMinD algorithms allow to precisely guide an autonomous mobile robot through a dense tree row solely using an RGB-D camera. A pear crop row in Gazebo is shown in the left picture, a real vineyard row on the right.}
    \label{fig:robots}
\end{figure}

Among the most researched uses are row-based crops, which account for about $75\%$ of all planted acres of agriculture in the United States \cite{Bigelow:263079}. The research involved in this scenario includes localization \cite{winterhalter2021localization}, path planning \cite{salvetti2023waypoint}, navigation \cite{man2020research, martini2022}, monitoring \cite{comba20192d}, harvesting \cite{harvesting}, spraying \cite{spraygrape}, and vegetative assessment \cite{zhang2020assessment, feng2020yield}. It can be especially difficult when line-of-sight obstructions or adverse weather prevent traditional localization techniques like GPS from achieving the required precision. This is evident in dense tree canopies, as demonstrated in Figure \ref{fig:robots}, which depicts a simulated pear orchard. 

Previous works have proposed position-agnostic vision-based navigation algorithms for row-based crops, as discussed in Section \ref{sec:related_works}. 

This work tackles a more challenging scenario in which dense canopies partially or totally cover the sky, and the GPS signal is very weak. We design a navigation algorithm based on semantic segmentation that exploits visual perception to estimate the center of the crop row and align the robot trajectory to it. The segmentation masks are predicted by a deep learning model designed for real-time efficiency and trained on realistic synthetic images. The proposed navigation algorithm improves on previous works being adaptive to different terrains and crops, including dense canopies. We conduct extensive experimentation in simulated and real environments for multiple crops. We compare our solution with previous state-of-the-art methodologies, demonstrating that the proposed navigation system is effective and adaptive to numerous scenarios.

\begin{figure*}[ht]
    \centering
    \includegraphics[width=\textwidth]{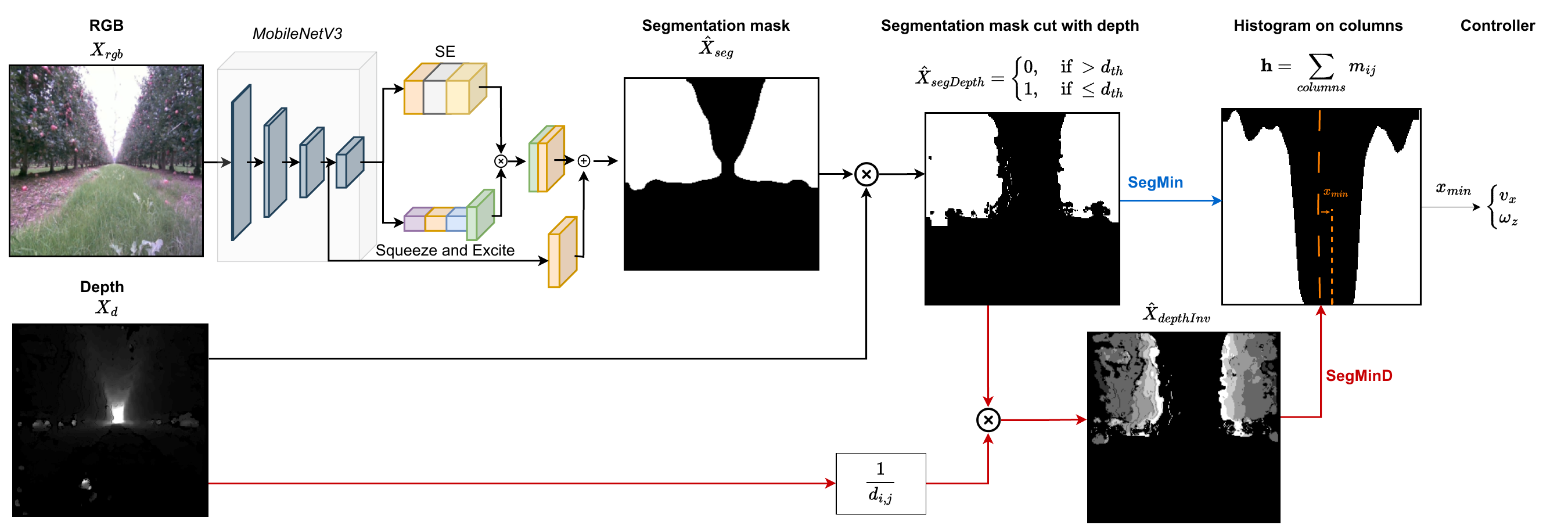}
    \caption{Scheme of the overall proposed navigation pipeline. The RGB image is fed into the segmentation network, thus the predicted segmentation mask $\hat{\textbf{X}}_{seg}^{t}$ is refined using the depth frame to obtain $\hat{\textbf{X}}_{segDepth}^{t}$. The \textcolor{RoyalBlue}{blue} arrow refers to the SegMin variant, and \textcolor{BrickRed}{red} arrows refer to the SegMinD variant to compute the sum histogram over the mask columns. Images are taken from navigation in the tall trees simulation world.}
    \label{fig:schema}
\end{figure*}

The main contributions of this work can be summarized as follows:
\begin{itemize}
    \item we present two variants of a novel approach for segmentation-based autonomous navigation in tall crops, designed to tackle challenging and previously uncovered scenarios;
    \item we test the resulting guidance algorithm on previously unseen plant rows scenarios such as high orchards trees and pergola vineyards.
    \item we train an efficient segmentation neural network with synthetic multi-crop data only, proving the generalization properties of the Deep Learning pipeline;
    \item we compare the new methods with state-of-the-art solutions on simulated and real vineyards, demonstrating an enhanced general and robust performance.
\end{itemize}

The next sections are organized as follows: Section \ref{sec:related_works} briefly describes previous approaches presented in the literature to tackle visual-based navigation in row-based crops. Section \ref{sec:methodology} presents the proposed deep-learning-based control system for vision-based position-agnostic autonomous navigation in row-based crops, from the segmentation model to the controller. Section \ref{sec:exp_results} describes the experimental setting and reports the main results for validating the proposed solution divided by sub-system. Finally, Section \ref{sec:conclusion} draws conclusive comments on the work and suggests interesting future directions.

\section{Related works}\label{sec:related_works}
The competitive advantage of GPS-free visual approaches has been partially investigated in the last years of service robotics research. Different computer vision and machine learning techniques have been applied so far to improve the performance of robot guidance.
A first vision-based approach was proposed in \cite{sharifi2015novel} using mean-shift clustering and the Hough transform to segment RGB images and generate the optimal central path. Later, \cite{RADCLIFFE2018165} achieved promising results using multispectral images and simply thresholding and filtering on the green channel. Recently, deep-learning approaches have been successfully applied to the task. \cite{huang2021endtoend, aghi2020autonomous} proposed a classification-based approach in which a model predicts the discrete action to perform. In contrast, \cite{aghi2020local, aghi2021deep} proposed a proportional controller to align the robot to the center of the row using heatmaps of the scene first and segmented images in the latest version. Finally, a different approach was introduced in \cite{martini2022} with an end-to-end controller based on deep reinforcement learning and depth images.
Although these systems proved effective in their testing scenarios, they have only been applied in simple crops where a full view of the sky favors both GPS receivers \cite{GPS_accuracy} and vision-based algorithms \cite{zaman2019cost}. 
Moreover, the increasing necessity of huge amounts of data to train Deep Learning models guided researchers to build wide and reliable synthetic datasets for crop semantic segmentation, as the one exploited in this work \cite{martini2023enhancing, angarano2023domain}.

\section{Methodology}\label{sec:methodology}
To navigate high-vegetation orchards and arboriculture fields, this work provides a real-time control algorithm with two variations, which enhances the method described in \cite{aghi2021deep}. The proposed method completely avoids the employ of GPS localization, which can be less accurate due to signal reflection and mitigation due to high and thick vegetation. 
Therefore, our algorithms consist of a straightforward operating principle, which exclusively employs RGB-D data and processes it to obtain effective position-agnostic navigation. It can be summarized in the following four steps:

\begin{enumerate}
    \item \label{enum:1} Semantic segmentation of the RGB frame, with the purpose of identifying the relevant plants in the camera's field of view.
    \item \label{enum:2} Addition of the depth data to the segmented frame to enhance the spatial understanding of the surrounding vegetation of the robot.
    \item \label{enum:3}Searching for the direction towards the end of the vegetation row, given the previous information.
    \item \label{enum:4} Generation of the velocity commands for the robot to follow the row.
\end{enumerate}

However, the two suggested approaches only vary in steps \ref{enum:2} and \ref{enum:3}, where they utilize depth frame data and generate the robot's desired direction. Conversely, the segmentation technique \ref{enum:1} and the command generation \ref{enum:4} are executed in a similar manner. A visual depiction of the proposed pipeline is illustrated in Figure \ref{fig:schema}.

The first step of the proposed algorithm, at each time instant $t$ consists in acquiring an RGB frame $\mathbf{X}_{rgb}^{t}$ and a depth frame $\mathbf{X}_d^t$, where $\mathbf{X}_{rgb}^{t} \in \mathbb{R}^{h \times w \times c}$ and $\mathbf{X}_{d}^{t} \in \mathbb{R}^{h \times w}$. In both cases, $h$ represents the frame height, $w$ represents the frame width, and $c$ is the number of channels. 
The RGB data received is subsequently inputted into a segmentation neural network model $H_{seg}$, yielding a binary segmentation mask that conveys the semantic information of the input frame.
\begin{equation}
    \hat{\textbf{X}}_{seg}^{t} = H \left(\textbf{X}_{rgb}^{t}\right)
    \label{eq:segmentation}
\end{equation}
where $\hat{\textbf{X}}_{seg}^{t}$ is the obtained segmentation mask.

Furthermore, the segmentation masks from the previous $N$ time instances, ranging from $t-N$ to $t$, are combined to enhance the robustness of the information.
\begin{equation}
    \hat{\textbf{X}}_{CumSeg}^{t} = \bigcup_{j = t-N}^{t} \hat{\textbf{X}}_{seg}^{j}
    \label{eq:cumulative-mask}
\end{equation}
where, $\hat{\textbf{X}}_{CumSeg}^{t}$ denotes the cumulative segmentation mask, and the symbol $\bigcup$ signifies the logical bitwise $OR$ operation applied to the last $N$ binary frames.

Moreover, the depth map $\mathbf{X}_{d}^{t}$ is employed to assess the segmented regions between the camera position and a specified depth threshold $d_{th}$. This process helps eliminate irrelevant information originating from distant vegetation, which has no bearing on controlling the robot's movement.
\begin{equation}
    \hat{\textbf{X}}_{segDepth\substack{i = 0,\dots,h\\ j = 0,\dots,w}}^{t} (i, j)
    = 
    \begin{cases}
        0, \textrm{ if } \hat{\textbf{X}}_{CumSeg (i, j)}^{t} \cdot \hat{\textbf{X}}^{t}_{d(i, j)} > d_{th}\\
        1, \textrm{ if } \hat{\textbf{X}}_{CumSeg (i, j)}^{t}\cdot \hat{\textbf{X}}^{t}_{d(i, j)} \leq d_{th}
    \end{cases}
    \label{eq:depth}
\end{equation}

\noindent where, $\hat{\textbf{X}}_{segDepth}$ represents the resultant intersection of the cumulative segmentation frame and the depth map, restricted to a distance threshold of $d_{th}$.

From this point forward, the proposed algorithm diverges into two variants, namely, \textit{SegMin} and \textit{SegMinD}, as elaborated in \ref{subsec:varA} and \ref{subsec:varB}, respectively.

\subsection{SegMin}
\label{subsec:varA}

The initial variant refines the methodology introduced in \cite{aghi2021deep}. Following the segmentation mask processing, a column-wise summation is executed, generating a histogram $\textbf{h} \in \mathbb{R}^{w}$ that characterizes the vegetation distribution along each column as in the following formula:
\begin{equation}
    \textbf{h}_{j} =  \sum_{i=1}^{h} {\hat{\textbf{X}}_{segDepth (i, j)}}
\end{equation}
where $i = 0, \dots, w$ is the index along the vertical direction of each frame column.

Subsequently, a moving average is applied to this histogram using a window of size $n$ to enhance robustness by smoothing values and mitigating punctual noise from previous passes. In an ideal scenario, the minimum value $x_{h}$ in this histogram corresponds to regions with minimal vegetation, effectively pinpointing the central path within the crop row.
If multiple global minima are identified, indicating areas with no detected vegetation, the mean of these points is calculated and considered as the global minimum. This approach ensures a more reliable identification of the continuation of the row, accommodating variations in the vegetation distribution.

\subsection{SegMinD}
\label{subsec:varB}
The second proposed methodology presents a variation of the earlier algorithm tailored specifically for wide rows featuring tall and dense canopies. In such scenarios, the initial algorithm might encounter challenges in determining a clear global minimum, as the consistent presence of vegetation above the robot complicates the interpretation. This variant addresses this issue by incorporating a multiplication operation between the previously processed segmentation mask and the normalized inverted depth data.
\begin{equation}
     \hat{\textbf{X}}_{depthInv}^{t} = \hat{\textbf{X}}_{segDepth}^{t}  \bigcap \left(1 - \dfrac{\textbf{X}_{d}^t}{d_{th}}\right)
     \label{eq:depth-inv}
\end{equation}

where, $ \hat{\textbf{X}}_{depthInv}^{t}$ is the outcome of an element-wise multiplication, denoted by $\bigcap$, involving the binary mask $\hat{\textbf{X}}_{segDepth}^{t}$ and the depth frame $\hat{\textbf{X}}_{d}^{t}$ that has been normalized over the depth threshold $d_{th}$. Similar to the previous scenario, a column-wise summation is executed to derive the array $\textbf{h}$, followed by a smoothing process using a moving average.

This introduced modification serves a crucial purpose by allowing elements closer to the robot to exert a more significant influence, thereby enhancing the algorithm's ability to discern the direction of the row.

\begin{figure}[H]
    \setlength{\fboxsep}{0pt}
    \centering
    \resizebox{0.85\columnwidth}{!}{%
    \begin{tabular}{c c c}
         RGB & SegMin & SegMinD  \\
         \framebox{\includegraphics[width=0.2\columnwidth]{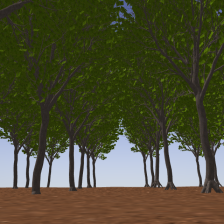}} & \framebox{\includegraphics[width=0.2\columnwidth]{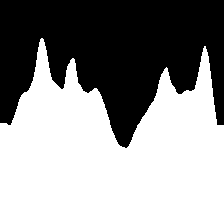}} & \framebox{\includegraphics[width=0.2\columnwidth]{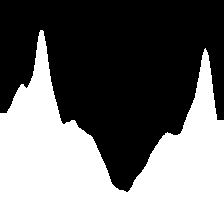}}
    \end{tabular}
    }
    \caption{Contrasting the histograms produced by the two distinct algorithms, considering the RGB frame on the right, reveals that SegMinD provides a more defined and less ambiguous global minimum point.}
    \label{fig:histos}
\end{figure}

The different sum histograms obtained with SegMin and SegMinD are directly compared in Figure \ref{fig:histos}, showing the sharper trend and the global minimum isolation obtained, including the depth values.

\subsection{Segmentation Network}\label{subsec:network}
\begin{figure}
    \centering
    \includegraphics[width=\textwidth]{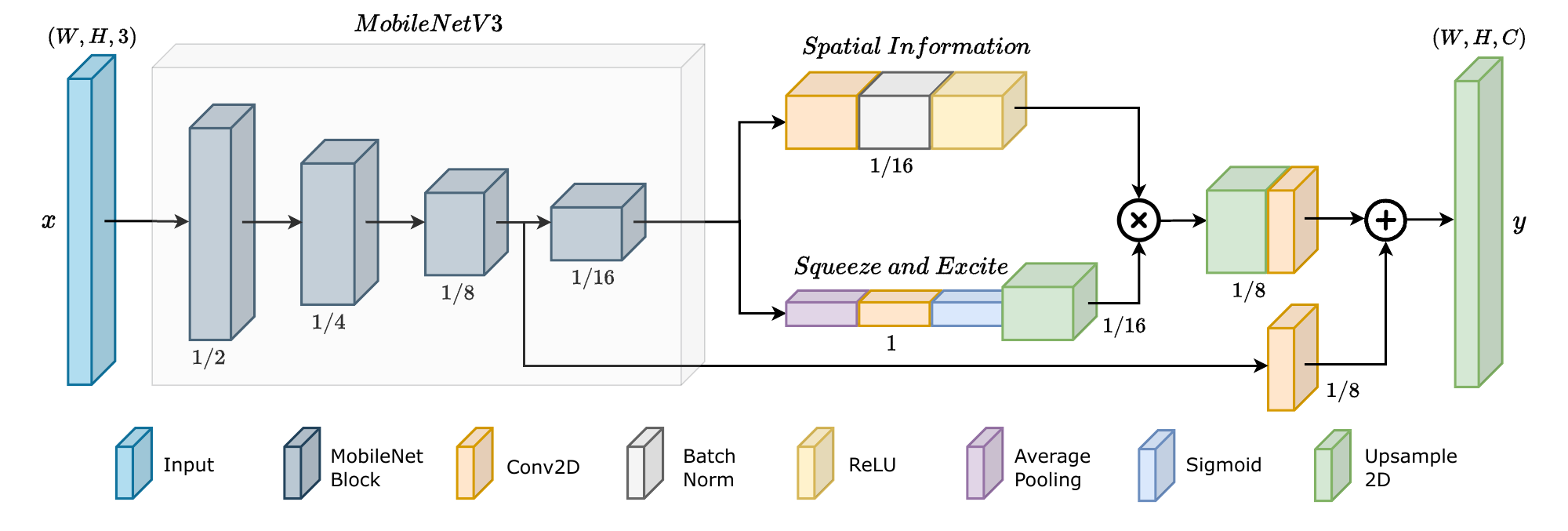}
    \caption{The Deep Neural Network utilized in this study features a backbone of MobileNetV3 and an LR-ASPP head, as detailed in \cite{howard2019searching}. The spatial scaling factor of the features in comparison to the input size is provided beneath each block.}
    \label{fig:architecture}
\end{figure}

A prior study on crop segmentation in real-world conditions provided the neural network design that was chosen \cite{aghi2021deep}. Figure \ref{fig:architecture} illustrates its entire architecture and its primary benefit is its ability to leverage rich contextual information from the image at a lower computing cost.

A MobileNetV3 backbone makes up the network's initial stage, which is designed to efficiently extract the visual features from the input image \cite{howard2019searching}. With squeeze-and-excitation attention sub-modules \cite{squeeze-and-excitation_hu_2018}, it is comprised of a series of inverted residual blocks \cite{sandler_mobilenetv2_2018}. They increase the amount of channel features while gradually decreasing the input image's spatial dimensions.

It is succeeded by a Lite R-ASPP (LR-ASPP) module \cite{chen_rethinking_2017}, an enhanced and condensed variant of the Atrous Spatial Pyramid Pooling module (R-ASPP) that upscales the extracted features via two parallel branches. The first lower the spatial dimension by $1/16$ by applying a Squeeze-and-Excite sub-module to the final layer of the backbone. To modify the number of channels $C$ to the output segmentation map, a channel attention weight matrix is produced, multiplied by the unpooled features, and then upsampled and fed through a convolutional layer. The second branch takes characteristics from an earlier stage of the backbone, which reduces the spatial dimension by $1/8$, and adds them to the output of the upsampling step, mixing lower-level and higher-level patterns in the data.

The network's input has a dimensionality equal to $W\times H\times 3$, while the segmented output is equal to $W\times H$.

Furthermore, the output values of the neural network are scaled between 0 and 1 using a sigmoid function, as this work primarily focuses on the semantic segmentation of plant rows.

The usual cross-entropy loss between the ground-truth label $y$ and the anticipated segmentation mask is used to train the DNN:

\begin{equation}
    L_\text{CE}(y,\hat y) = -\sum_{i=1}^{N} y_i\cdot log(\hat y_i)
\end{equation}
\noindent which for binary segmentation becomes a simple binary cross-entropy loss. 

During both the validation and testing phases, the DNN performance is evaluated through an intersection over unit (IoU) metrics:

\begin{equation}
    mIoU(\theta) = \frac{1}{N}\sum_{i=0}^{N} \left(1 - \frac{\hat{X}_{seg}^{i} \cap X_{seg}^{i}}{\hat{X}_{seg}^{i} \cup X_{seg}^{i}}\right)
\end{equation}

where $X_{seg}^i$ is the ground truth mask, $\hat{X}_{seg}^i$ is a predicted segmentation mask, and $\theta$ is the vector representing the network parameters. Since there are only plants as the target class of interest, $N$ in the IoU computation always equals 1.
The model is trained on the AgriSeg synthetic dataset \cite{martini2023enhancing}\footnote{\url{https://pic4ser.polito.it/AgriSeg}}. Further details on the training strategy and hyperparameters are provided in Section \ref{sec:exp_results}.

\subsection{Robot heading control}
The goal of the controller pipeline is to maintain the mobile platform at the center of the row, which, in this study, is equated to aligning the row center with the middle of the camera frame. Consequently, following the definition in the preceding step, the minimum of the histogram should be positioned at the center of the frame width. The distance $d$ from the frame center to the minimum is defined as:
\begin{equation}
    d = x_{h} - \dfrac{w}{2}
\end{equation}

The generation of linear and angular velocities is accomplished using custom functions, mirroring the approach employed in \cite{cerrato2021deeplearning}.
\begin{equation}
    v_{x} = v_{x, max} \left[1 - \dfrac{d^{2}}{\left(\frac{w}{2}\right)^{2}} \right]
\end{equation}

\begin{equation}
    \omega_{z} = -k_{\omega_z}\cdot \omega_{z, max}\cdot \dfrac{d^2}{w^2}
    \label{eq:ang_vel}
\end{equation}

where, $v_{x, max}$ and $\omega_{z, max}$ represent the maximum attainable linear and angular velocities, and $k_{\omega_z}$ serves as the angular gain controlling the response speed. To mitigate abrupt changes in the robot's motion, the ultimate velocity commands $\bar{v}{x}$ and $\bar{\omega}{z}$ undergo smoothing using an Exponential Moving Average (EMA), expressed as:
\begin{equation}
    \begin{bmatrix}
        \bar{v}_{x}^{t} \\ \bar{\omega}_{z}^{t}
    \end{bmatrix}
     = (1 - \lambda) \begin{bmatrix}
        \bar{v}_{x}^{t-1} \\ \bar{\omega}_{z}^{t-1}
    \end{bmatrix}
    + \lambda \begin{bmatrix}
        {v}_{x}^{t} \\ \omega_{z}^{t}
    \end{bmatrix}
\end{equation}

where, $t$ represents the time step, and $\lambda$ stands for a selected weight.

\section{Experiments and Results}
\label{sec:exp_results}

\begin{figure}[]
\centering
\begin{tabular}{cc}
    \includegraphics[width=0.8\columnwidth]{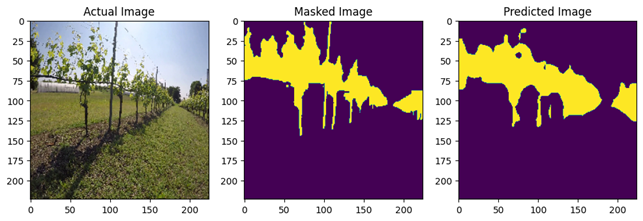} \\
    \includegraphics[width=0.8\columnwidth]{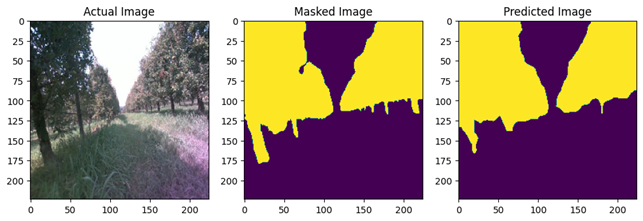} \\
    \includegraphics[width=0.8\columnwidth]{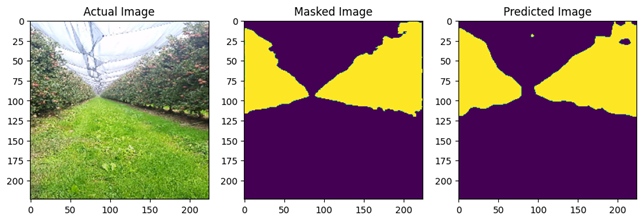}
\end{tabular}
\caption{Test of semantic segmentation DNN on real-world test samples from vineyard (top), pear trees (middle) and apple trees (bottom) fields. For each crop, RGB input image (left), ground truth mask (center) and the predicted mask (right) are reported.}
\label{fig:seg_test_real}
\end{figure}

\subsection{Segmentation Network Training and Evaluation} \label{subsec:seg_test}
We train the crop segmentation model using a subset of the AgriSeg synthetic segmentation dataset \cite{martini2023enhancing, angarano2023domain}. In particular, for the pear trees and apple trees, we train on generic tree datasets in addition to pear and apples; for vineyards, we train on vineyard and pergola vineyards (note that the testing environments are different from the ones from which the training samples are generated). Only $100$ miscellaneous real images of different crop types are added to the training dataset in all the cases. Thanks to the high-quality rendering of the AgriSeg dataset, this small amount of real images is sufficient to reach general and robust performance in real-world conditions. In both cases, the model is trained for 30 epochs with Adam optimizer and learning rate $3\times10^{-4}$. We apply data augmentation by randomly applying cropping, flipping, greyscaling, and random jitter to the images. Our experimentation code is developed in Python 3 using TensorFlow as the deep learning framework. We train models starting from ImageNet pretrained weights, so the input size is fixed to (224 × 224). All the training runs are performed on a single Nvidia RTX 3090 graphic card.

\begin{table}[ht]
    \centering
    \caption{Semantic Segmentation results on real images in different crop fields. For each crop, a model has been trained on synthetic data, only using 100 additional real images containing miscellaneous crops different from the test set.}
    \label{tab:seg_real_results}
    \begin{tabular}{lccc}
      
    \toprule
       \textbf{Model}  & \textbf{Real Test IoU} & \textbf{Train  Data} &\textbf{Real Test Data} \\ \midrule

       Vineyard  & \textbf{0.6950} & 13840 & 500  \\
       Apples  & \textbf{0.8398}  & 15280 & 210 \\
       Pear  & \textbf{0.8778} & 7980 & 140  \\ 
       \bottomrule
    \end{tabular}
\end{table}

Table \ref{tab:seg_real_results} reports the results obtained testing the trained segmentation DNN on real images in terms of Intersection over Union (IoU). Figure \ref{fig:seg_test_real} also shows some qualitative results on sample images collected on the field during the test campaign.

\subsection{Simulation Environment}
The proposed control algorithm underwent testing using the Gazebo simulation software\footnote{\url{https://gazebosim.org}}. Gazebo was chosen due to its compatibility with ROS 2 and its ability to integrate plugins simulating sensors, including cameras. A Clearpath Jackal model was employed to evaluate the algorithm's performance. The URDF file from Clearpath Robotics, containing comprehensive information about the robot's mechanical structure and joints, was utilized. In the simulation, an Intel Realsense D435i plugin was employed, placed 20 cm in front of the robot's center, and tilted upward by $15^{\circ}$: this configuration enhanced the camera's visibility of the upper branches of trees.

The assessment of the navigation algorithm took place in four customized simulation environments, each designed to mirror distinct agricultural scenarios. These environments included a conventional vineyard, a pergola vineyard characterized by elevated vine poles and shoots above the rows, a pear field populated with small-sized trees, and a high-tree field where the canopies interweave above the rows. Each simulated field features varied terrains, replicating the irregularities found in real-world landscapes. Comprehensive measurements for each simulation world can be found in Table \ref{tab:mondi}.

In the experimental phase of this study, we adopted frame dimensions of $(h, w) = (224, 224)$, matching the input and output sizes of the neural network model, with a channel count of $c = 3$. The maximum linear velocity was set to $v_{x, max} = 0.5 m/s$, and the maximum angular velocity was capped at $\omega_{z, max} = 1 rad/s$. The angular velocity gain, denoted as $\omega_{z, gain}$, was fixed at $0.01$, and the Exponential Moving Average (EMA) buffer size was set to 3. Additionally, the depth threshold was adjusted based on the specific characteristics of different crops.

Specifically, it has been empirically set at 5 m for vineyards, raised to 8 m for pear trees and pergola vineyards, and further increased to 10 m for tall trees, taking into account the average distance from the rows in various fields.

\begin{figure}[]
\setlength{\fboxsep}{0pt}
\centering
\resizebox{0.9\columnwidth}{!}{%
\begin{tabular}{cccc}
& RGB & Masked Depth & Histogram\\ 
\vspace{3pt}
(a) & \framebox{\includegraphics[width=.15\columnwidth, valign=m]{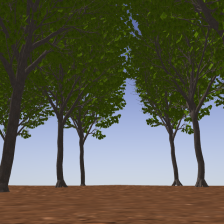}} & \framebox{\includegraphics[width=.15\columnwidth, valign=m]{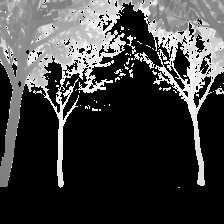}} & \framebox{\includegraphics[width=.15\columnwidth, valign=m]{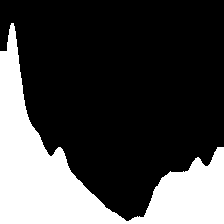}}\\ 
\vspace{3pt}
(b) & \framebox{\includegraphics[width=.15\columnwidth, valign=m]{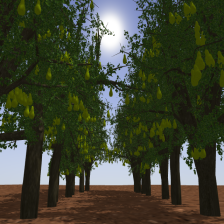}} & \framebox{\includegraphics[width=.15\columnwidth, valign=m]{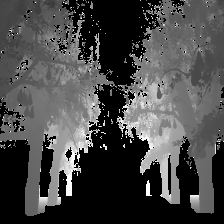}} & \framebox{\includegraphics[width=.15\columnwidth, valign=m]{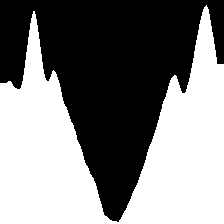}}\\
\vspace{3pt}
(c) & \framebox{\includegraphics[width=.15\columnwidth, valign=m]{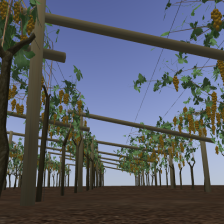}} & \framebox{\includegraphics[width=.15\columnwidth, valign=m]{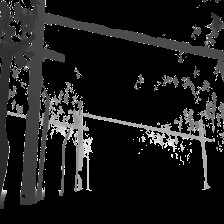}} & \framebox{\includegraphics[width=.15\columnwidth, valign=m]{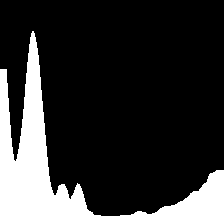}}\\
\vspace{3pt}
    (d) & \framebox{\includegraphics[width=.15\columnwidth, valign=m]{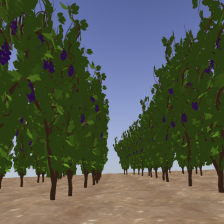}} & \framebox{\includegraphics[width=.15\columnwidth, valign=m]{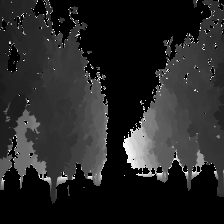}} & \framebox{\includegraphics[width=.15\columnwidth, valign=m]{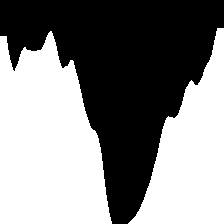}}\\
\end{tabular}
}
\caption{Sample outputs of the proposed SegMinD algorithm for High Trees (a), Pear Trees (b), Pergola Vineyard (c), and Vineyard (d). Predicted segmentation masks are refined cutting values exceeding a depth threshold. The sum over mask columns provides the histograms used to identify the center of the row as its global minimum.}
\end{figure}

\begin{table}[t]
\centering
\caption{
Dimensions of various simulated crops indicate the average values for the distance between rows, the spacing between plants within a row, and the heights of the plants.}
\begin{tabular}{lccc}
\toprule
\textbf{Gazebo worlds}    & \textbf{Rows distance {[}m{]}} & \textbf{Plant distance  {[}m{]}} & \textbf{Height {[}m{]}} \\ \midrule
Common vineyard  & 1.8                         & 1.3                             & 2.0               \\
Pergola vineyard & 6.0                           & 1.5                             & 2.9             \\
Pear field       & 2.0                           & 1.0                               & 2.9            \\
High trees field & 7.0                           & 5.0                               & 12.5            \\ \bottomrule
\end{tabular}%

\label{tab:mondi}
\end{table}

\begin{figure}[H]
    \centering
    \setlength{\fboxsep}{0pt}
    \centering
    \resizebox{0.9\columnwidth}{!}{%
    \begin{tabular}{c c}
    \includegraphics[width=0.5\columnwidth]{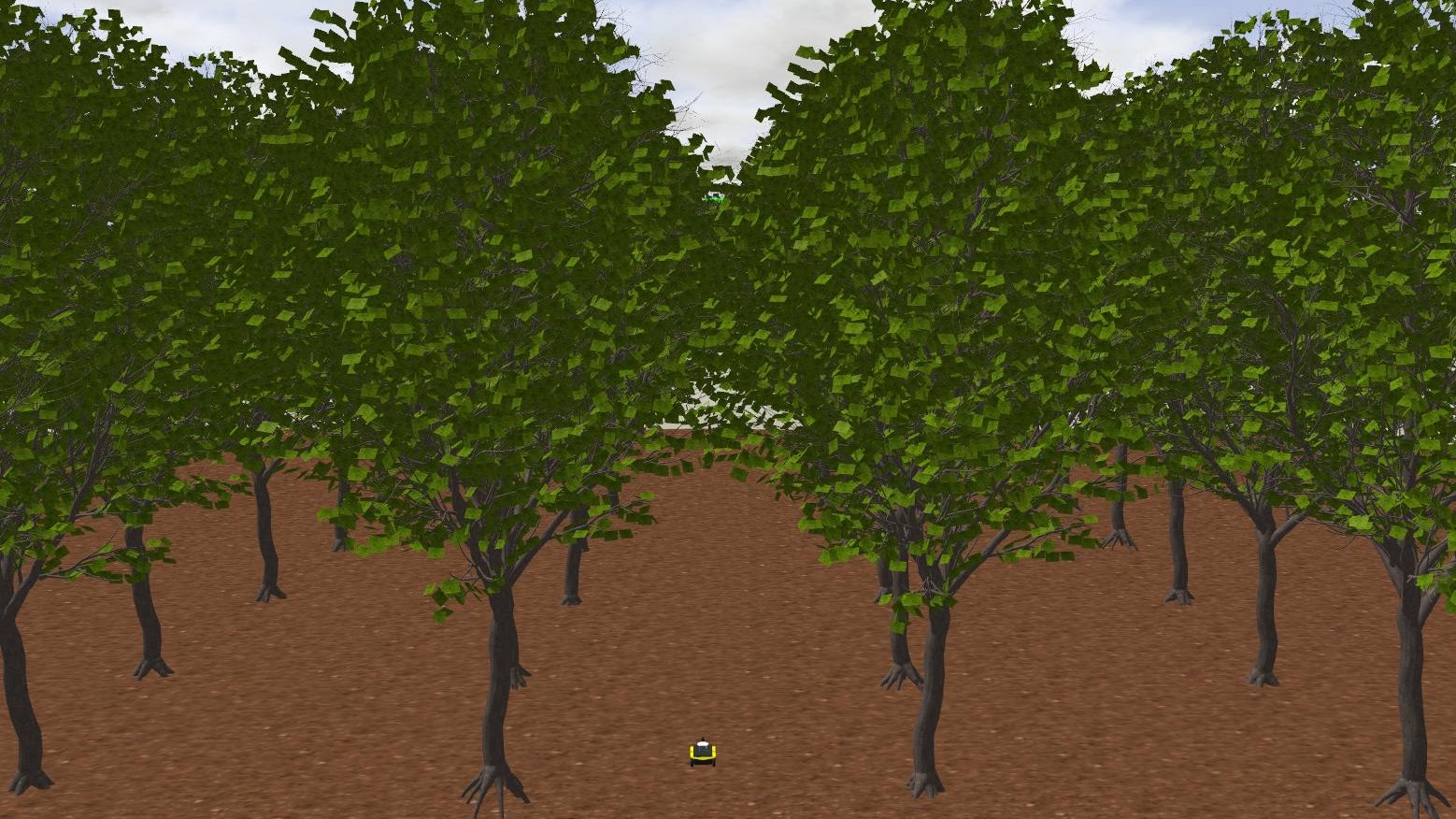} & \includegraphics[width=0.5\columnwidth]{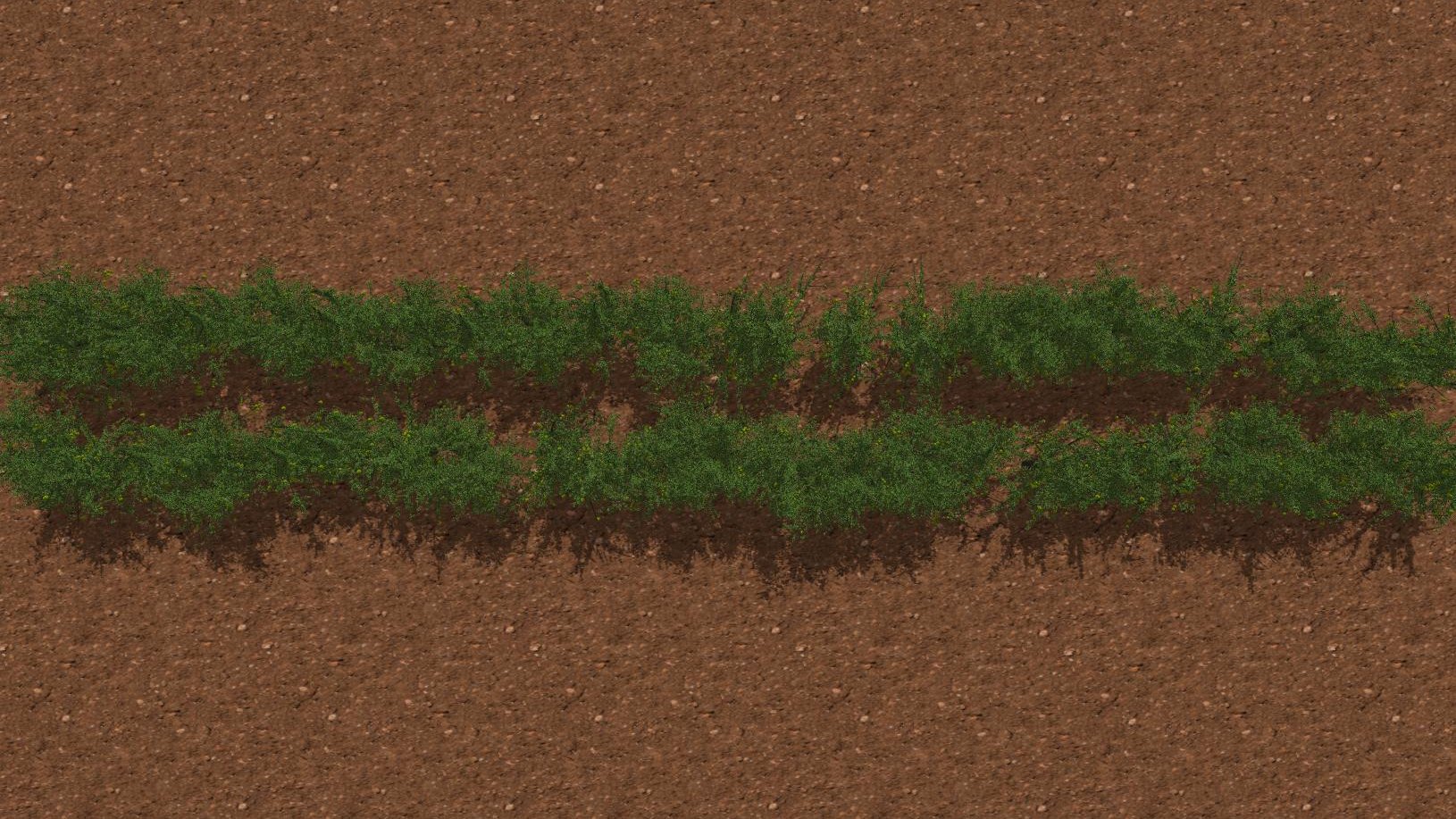} \\
    (a) & (b)\\
    \includegraphics[width=0.5\columnwidth]{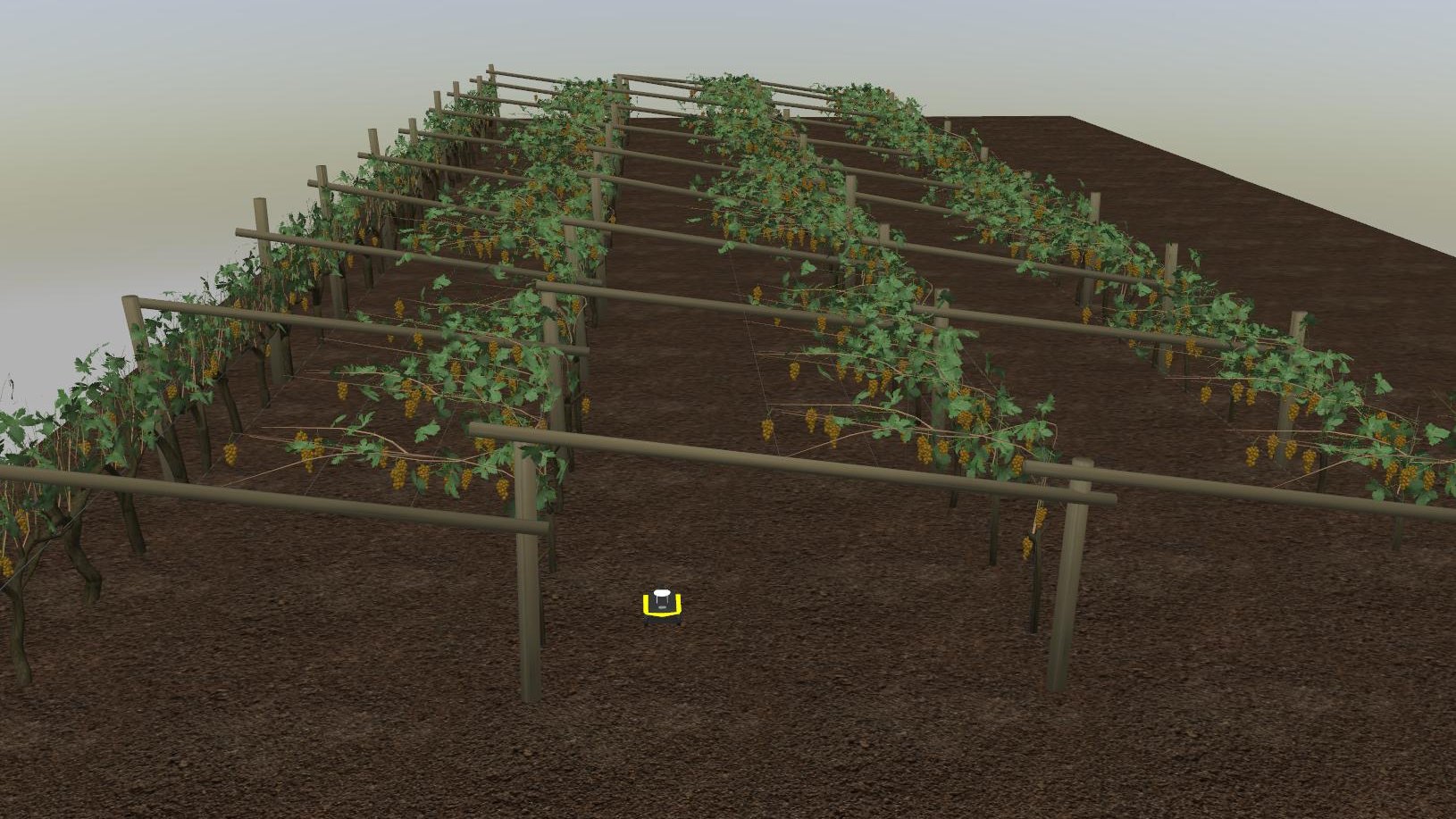} & \includegraphics[width=0.5\columnwidth]{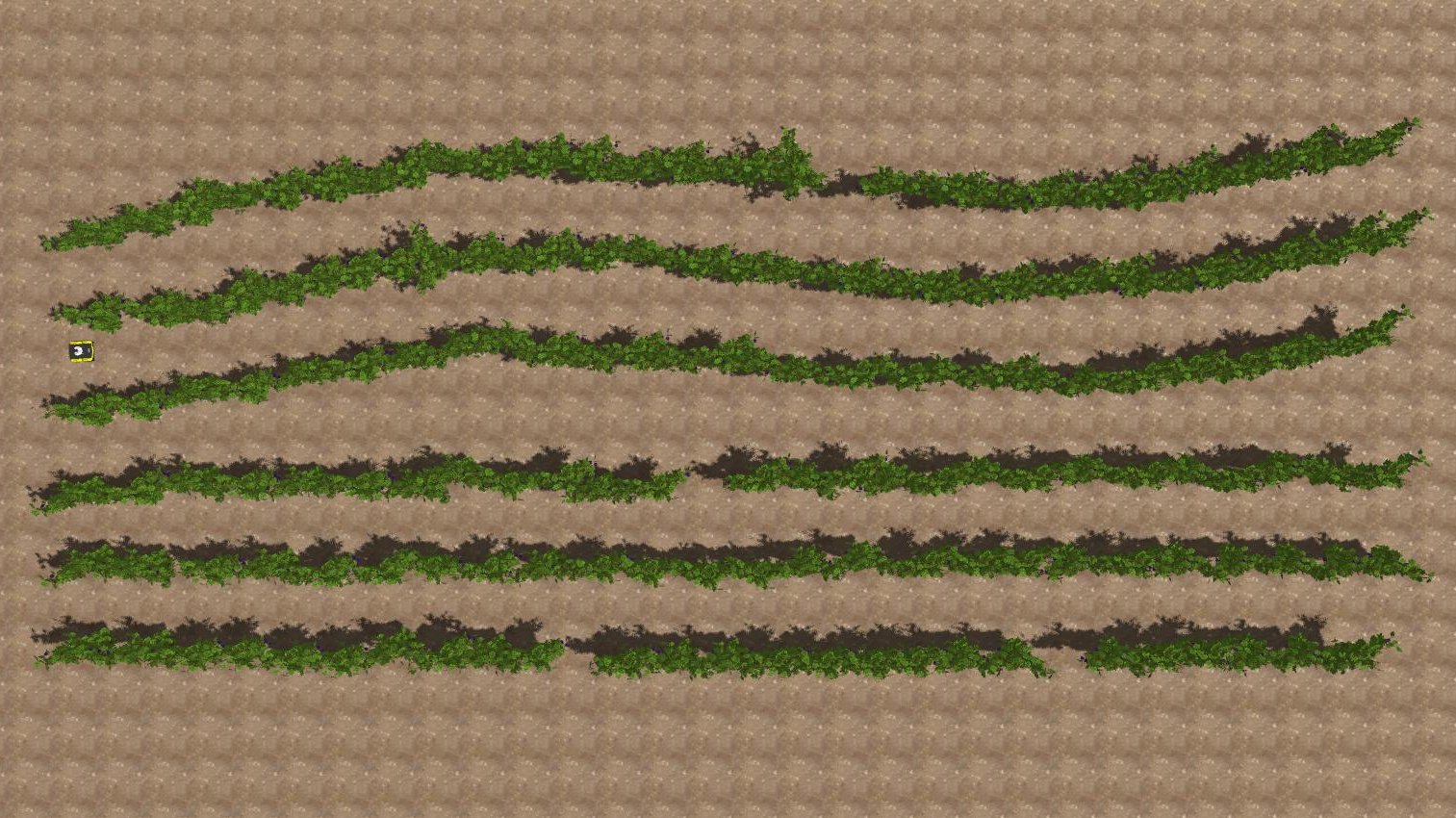} \\
    (c) & (d)\\
    \end{tabular}
    }
    \caption{Gazebo simulated environments were employed to assess the SegMin approach in various crop rows of significance, including wide rows with high trees (a), a slender row of pear trees (b), an asymmetric pergola vineyard with irregular rows (c), and both straight and curved vineyard rows (d). For the latter scenario, the evaluations were conducted in both the second row from the top and the second row from the bottom.}
    \label{fig:mondi}
\end{figure}

\subsection{Navigation Results in Simulation} 
\label{subsec:sim_test}

\begin{table}[t]
\centering
\caption{
Navigation outcomes across diverse test fields were assessed using the SegMin, SegMinD, and the SegZeros segmentation-based algorithms. The evaluation employed metrics to gauge the efficacy of navigation, including clearance time, and assessed precision through Mean Absolute Error (MAE) and Mean Squared Error (MSE) by comparing the obtained path with the ground truth. Additionally, kinematic information about the robot's navigation was captured through the cumulative heading average $\gamma [rad]$, mean linear velocity $v_{avg} [m/s]$, and the standard deviation of angular velocity $\omega_{stddev} [rad/s]$. Notably, SegZeros proved impractical for scenarios involving tall trees, pear trees, and pergola vineyards.}
\label{tab:test_results}
\resizebox{\columnwidth}{!}{
\begin{tabular}{@{}llccccccc@{}}
\toprule
\textbf{Test Field} &
  \textbf{Method} &
  \textbf{Clearance {[}s{]}} &
  \textbf{MAE {[}m{]}} &
  \textbf{MSE {[}m{]}} &
  \textbf{Cum. $\pmb{\gamma_{avg}}$ [rad]} &
  \textbf{$\pmb{v_{avg}}$ [m/s]} &
  $\pmb{\omega_{std dev}}$ \textbf{[rad/s]} \\ \midrule
\textbf{High Trees} & SegMin & \textbf{40.41 ± 0.12} & 0.27 ± 0.01 & 0.08 ± 0.00 & 0.08 ± 0.00 & 0.49 ± 0.00 & 0.05 ± 0.00 \\
                  & SegMinD & 40.44 ± 0.51 & \textbf{0.17 ± 0.01} & \textbf{0.04 ± 0.00} & 0.05 ± 0.00 & 0.48 ± 0.01 & 0.06 ± 0.02 \\\midrule
\textbf{Pear Trees} & SegMin& \textbf{42.06 ± 1.23} & 0.03 ± 0.01 & \textbf{0.00 ± 0.00} & 0.01 ± 0.00 & 0.48 ± 0.00 & 0.11 ± 0.05 \\
                  & SegMinD & 42.26 ± 1.91 &\textbf{0.03 ± 0.02} & \textbf{0.00 ± 0.00} & 0.02 ± 0.00 & 0.48 ± 0.01 & 0.03 ± 0.00 \\\midrule
\textbf{Pergola   Vine.} & SegMin & \textbf{40.86 ± 0.39} & \textbf{0.08 ± 0.01} & \textbf{0.01 ± 0.00} & 0.03 ± 0.02 & 0.48 ± 0.00 & 0.17 ± 0.02 \\
                  & SegMinD & 41.14 ± 0.33 & 0.10 ± 0.05 & 0.01 ± 0.01 & 0.03 ± 0.01 & 0.48 ± 0.00 & 0.20 ± 0.03 \\\midrule
\textbf{Straight   Vine.} & SegMin & \textbf{50.51 ± 0.31} & \textbf{0.11 ± 0.00} & \textbf{0.01 ± 0.00} & 0.03 ± 0.00 & 0.49 ± 0.00 & 0.08 ± 0.01 \\
                  & SegMinD & 50.63 ± 0.28 & 0.11 ± 0.01 & 0.02 ± 0.00 & 0.03 ± 0.01 & 0.49 ± 0.00 & 0.09 ± 0.01 \\
                  & SegZeros & 53.69 ± 1.03 & 0.14 ± 0.03 & 0.02 ± 0.01 & 0.03 ± 0.0 & 0.46 ± 0.01 & 0.09 ± 0.01 \\\midrule
\textbf{Curved   Vine.} & SegMin & 53.32 ± 0.25 & 0.12 ± 0.01 & 0.02 ± 0.00 & 0.04 ± 0.01 & 0.49 ± 0.00 & 0.09 ± 0.02 \\
                  & SegMinD & \textbf{51.44 ± 1.03} & \textbf{0.09 ± 0.01} & \textbf{0.01 ± 0.00} & 0.01 ± 0.00 & 0.48 ± 0.01 & 0.06 ± 0.01 \\
                  & SegZeros & 71.05 ± 27.13 & 0.11 ± 0.04 & 0.02 ± 0.01 & 0.05 ± 0.01 & 0.40 ± 0.13 & 0.11 ± 0.04 \\ 
                  \bottomrule
\end{tabular}
}
\end{table}

\begin{figure}[]
    \centering
    \includegraphics[width=0.85\columnwidth]{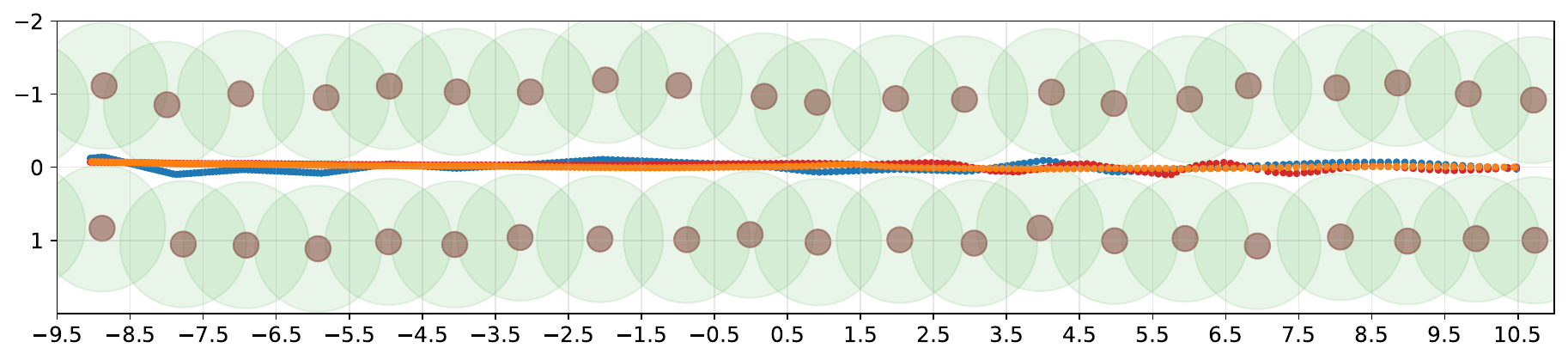}
    \includegraphics[width=0.85\columnwidth]{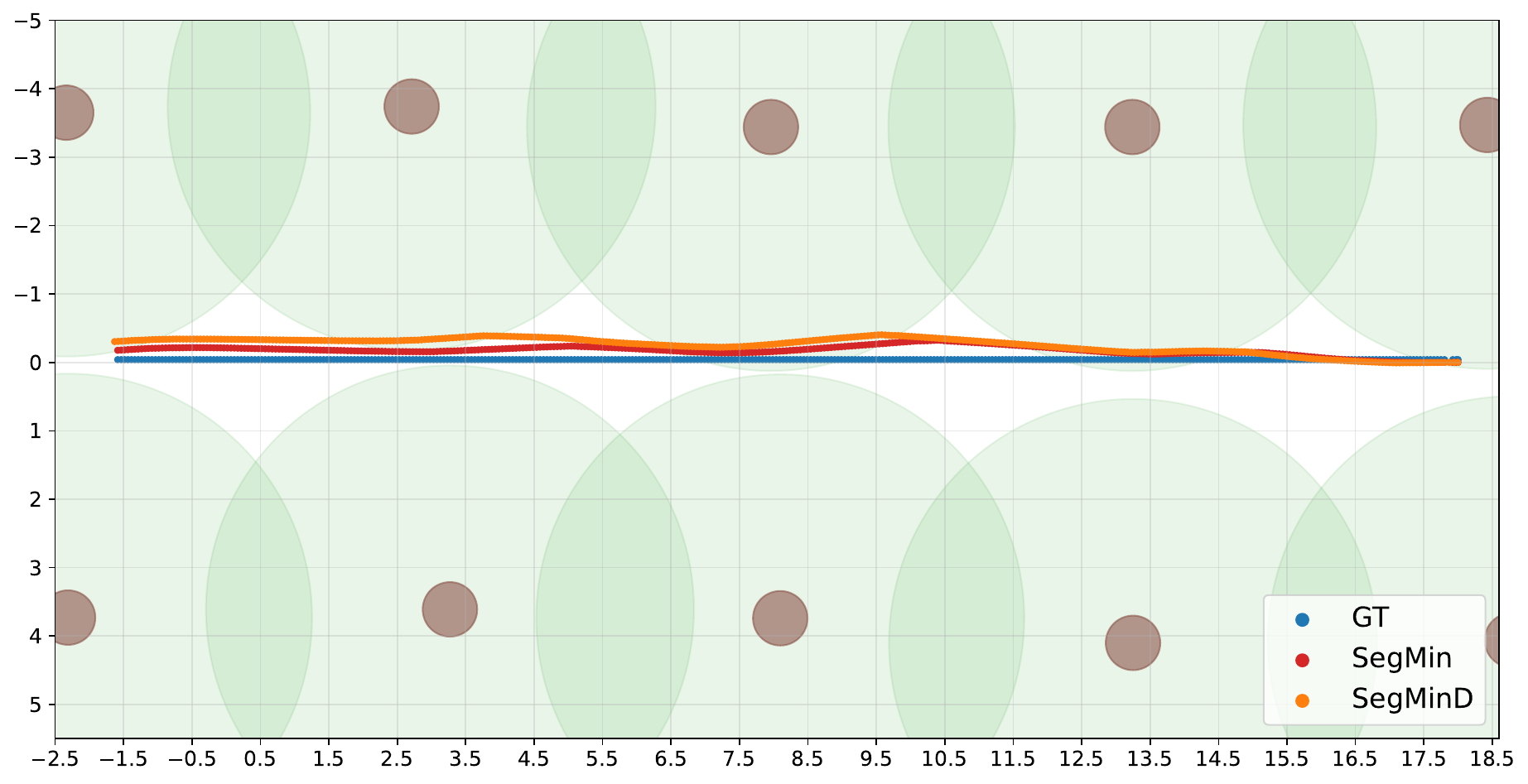}
    \includegraphics[width=0.85\columnwidth]{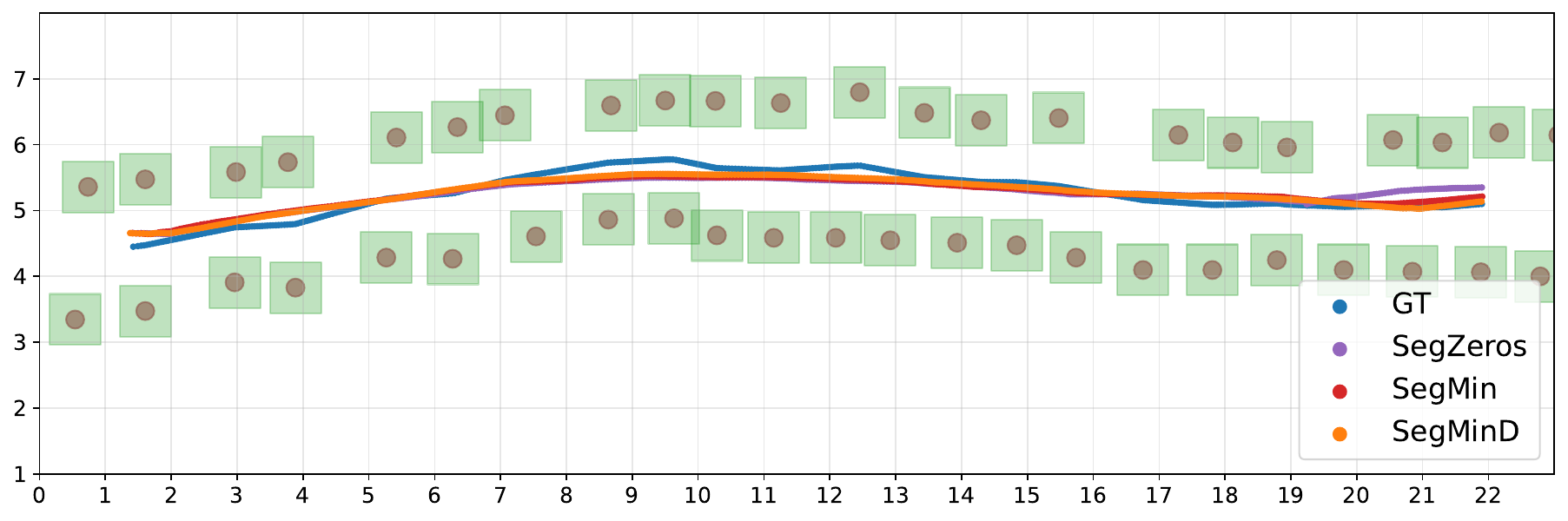}
    \caption{Trajectories comparison between our proposed algorithms (SegMin and SegMinD) and the ground truth central path (GT): Pears (top), High Trees (center), Curved Vineyard (bottom). In the last graph, the trajectory generated with the SegZeros algorithm is also reported for comparison.}
    \label{fig:trajectories}
\end{figure}

The comprehensive evaluation of the SegMin navigation pipeline and its variant, SegMinD, took place in realistic crop fields within a simulation environment, employing pertinent metrics for visual-based control without the need for precise robot localization, aligning with methodologies from prior studies \cite{aghi2021deep, martini2022}. The camera frames were published at a frequency of 30 Hz, with inference conducted at 20 Hz and velocity commands from controllers published at 5 Hz. The evaluation utilized the testing package from the open-source PIC4rl-gym\footnote{\url{https://github.com/PIC4SeR/PIC4rl\_gym}} in Gazebo \cite{martini2022pic4rl}.
The chosen metrics aimed to assess the navigation effectiveness, measured by clearance time and precision, involving a quantitative comparison of obtained trajectories with a ground truth trajectory using Mean Absolute Error (MAE) and Mean Squared Error (MSE). Ground truth trajectories were computed by averaging interpolated poses of plants within rows. In the case of an asymmetric pergola vineyard, a row referred to the portion without vegetation on top, as depicted in Figure \ref{fig:mondi} (c).
The algorithms' response to terrain irregularities and row geometries was also studied, encompassing significant kinematic information about the robot. The evaluation considered the cumulative heading average $\gamma [rad]$ along the path, mean linear velocity $v_{avg} [m/s]$, and standard deviation of angular velocity $\omega_{stddev} [rad/s]$. These metrics provided insights into how well the algorithms maintained the robot's correct orientation, with the mean value of $\omega$ consistently approaching zero due to successive orientation corrections.

The complete set of results is outlined in Table \ref{tab:test_results}. Each metric is accompanied by both the average value and standard deviation, reflecting the repetition of experiments in three runs on a 20 m long track within each crop row. The proposed method effectively addresses the challenge of guiding the robot through rows of trees with dense canopies, such as high trees and pears, even in the absence of a localization system. It also demonstrates proficiency in unique scenarios, like navigating through pergola vineyards.
The presence of plant branches and wooden supports poses a challenge for existing segmentation-based solutions. These solutions, built on the assumption of identifying a clear passage by focusing solely on zeros in the binary segmentation mask \cite{aghi2021deep}, encounter limitations in our tested scenarios. In our result comparisons, we term this prior method as SegZeros, utilizing the same segmentation neural network for assessment.

The SegMin methodology, based on histogram minimum search, proves to be a resilient solution for guiding the robot through tree rows. The incorporation of depth inverse values as a weighting function in SegMinD enhances the algorithm's precision, particularly in navigating through challenging scenarios like wide rows (high trees) and curved rows (curved vineyard). Furthermore, these innovative methods exhibit competitive performance even in standard crop rows, where a clear passage to the end of the row is discernible in the mask without canopy interference.
Compared to the previous segmentation-based baseline method, the histogram minimum approach significantly reduces navigation time and enhances trajectory precision in both straight and curved vineyard rows. On the other hand, the search for plant-free zero clusters in the map proves to be less robust and efficient, leading to undesired stops and an overall slower and more oscillating behavior during navigation. Additionally, the standard deviation of angular velocity aligns with the results obtained, being smaller in cases where the trajectory is more accurate, while the cumulative heading exhibits larger values when the algorithms demonstrate increased reactivity.

However, the trajectories generated by the SegMin, SegMinD, and SegZeros algorithms are visually depicted in Figure \ref{fig:trajectories} within representative scenarios. These scenarios include a cluttered, narrow row featuring small pear trees, a wide row with high trees, and curved vineyards where the state-of-the-art SegZeros method is applied.

\subsection{Navigation Test on the field} 
\label{subsec:real_test}

The overall navigation pipeline of SegMin and its variant SegMinD are tested in real crop fields, evaluating the results with relevant metrics for visual-based control without precise localization of the robot, as done in previous works \cite{aghi2021deep, martini2022}. The robotic platform employed to perform the tests is a Clearpath Husky UGV equipped with a LiDAR Velodyne Puck VLP-16, an RGBD camera Realsense D455, an AHRS Microstrain 3DM-GX5 and a Mini-ITX computer with an Intel Core i7 processor and 16 GB of memory. The camera frames were captured at a rate of 30 Hz, inference was performed at 20 Hz, and velocity commands were published at 5 Hz. In this section's experiments, ground truth trajectory was unavailable due to the complexity and demanding nature of measurement, which requires sophisticated instruments for sufficient accuracy. Instead, the lateral displacement of the rover within the row was determined using point clouds from the LiDAR. Points were clustered to separate the two rows, then fitted by a straight line, followed by computing the shortest distance from the plants to the origin, i.e., the center of the robot, where the sensor is mounted, for both lanes. The AHRS measured the rover's heading and compared it with the average heading of the row obtained from satellite images, considering the tested rows are straight.

\begin{table}[]
    \centering
    \caption{Navigation results of the real-world testing of the algorithms. SegMinD has been tested on apple and pear orchards. A comparison has been performed between the algorithms SegZeros, SegMin and SegMinD in a vineyard row choosing the same parameters: depth threshold set to 8.0 m and pixel-wise confidence to 0.7.}
    \label{tab:algo-comp}
    \fontsize{10pt}{10pt}\selectfont
    \begin{tabular}{@{}llcccc@{}}
        \toprule
         & \textbf{Algorithm} & \textbf{MAE [m]} & \textbf{RMSE [m]} & \textbf{Cum. $\pmb{\gamma_{avg}}$ [rad]} & \textbf{$\omega$ STD [rad/s]} \\ \midrule
        \multirow{2}{*}{\textbf{Apple Trees}} & SegMin & 0.16708 & 0.18841 & -0.03278 & 0.04112 \\
         & SegMinD & 0.07208 & 0.09110 & 0.10709 & 0.06881 \\ \midrule
        \multirow{2}{*}{\textbf{Pear Trees}} & SegMin & 0.46540 & 0.47309 & 0.03066 & 0.12924 \\
         & SegMinD & 0.28435 & 0.29792 & 0.03004 & 0.13074 \\  \midrule
        \multirow{3}{*}{\textbf{Vineyard}} & SegZeros & 0.16669 & 0.17011 & -0.01282 & 0.02675 \\
         & SegMin & 0.23045 & 0.24193 & -0.03372 & 0.11451 \\
         & SegMinD & 0.16007 & 0.17093 & -0.11478 & 0.10127 \\ \bottomrule
    \end{tabular}
\end{table}

\begin{figure}[]
    \centering
    \includegraphics[width=.9\linewidth]{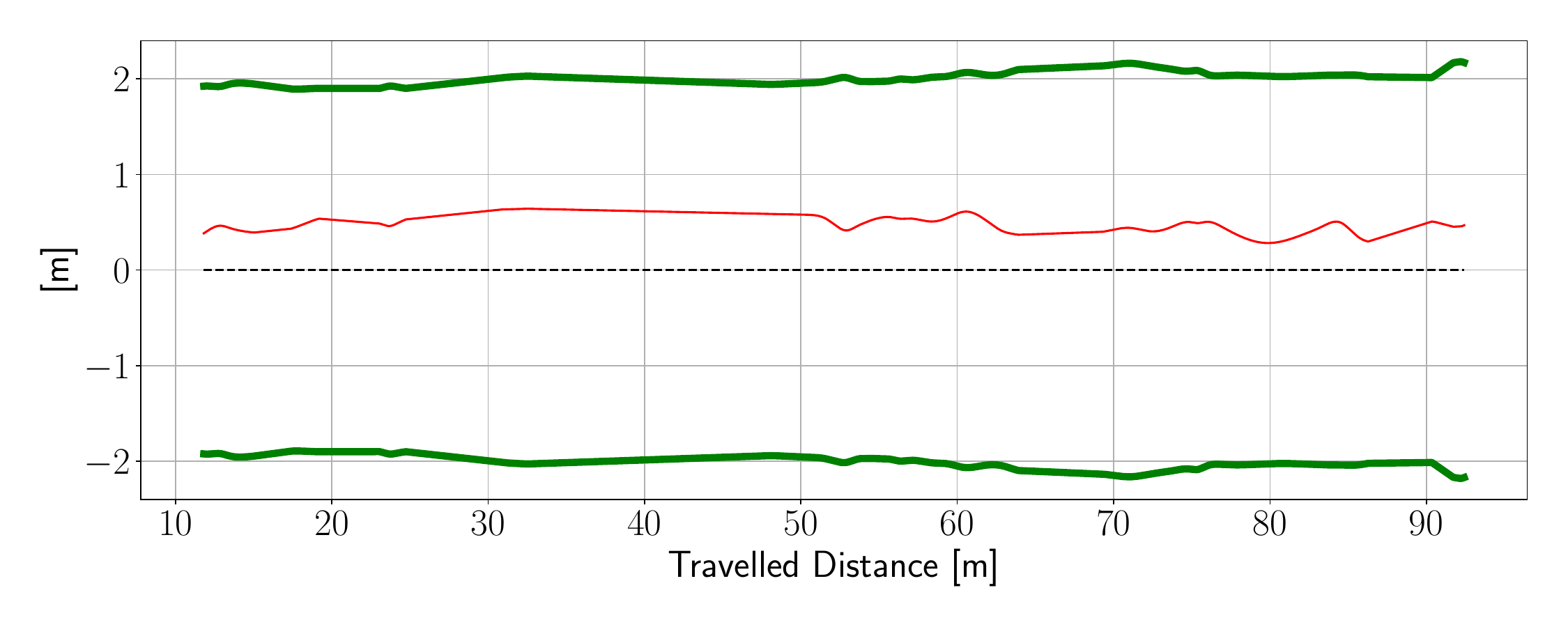}
    \includegraphics[width=.9\linewidth]{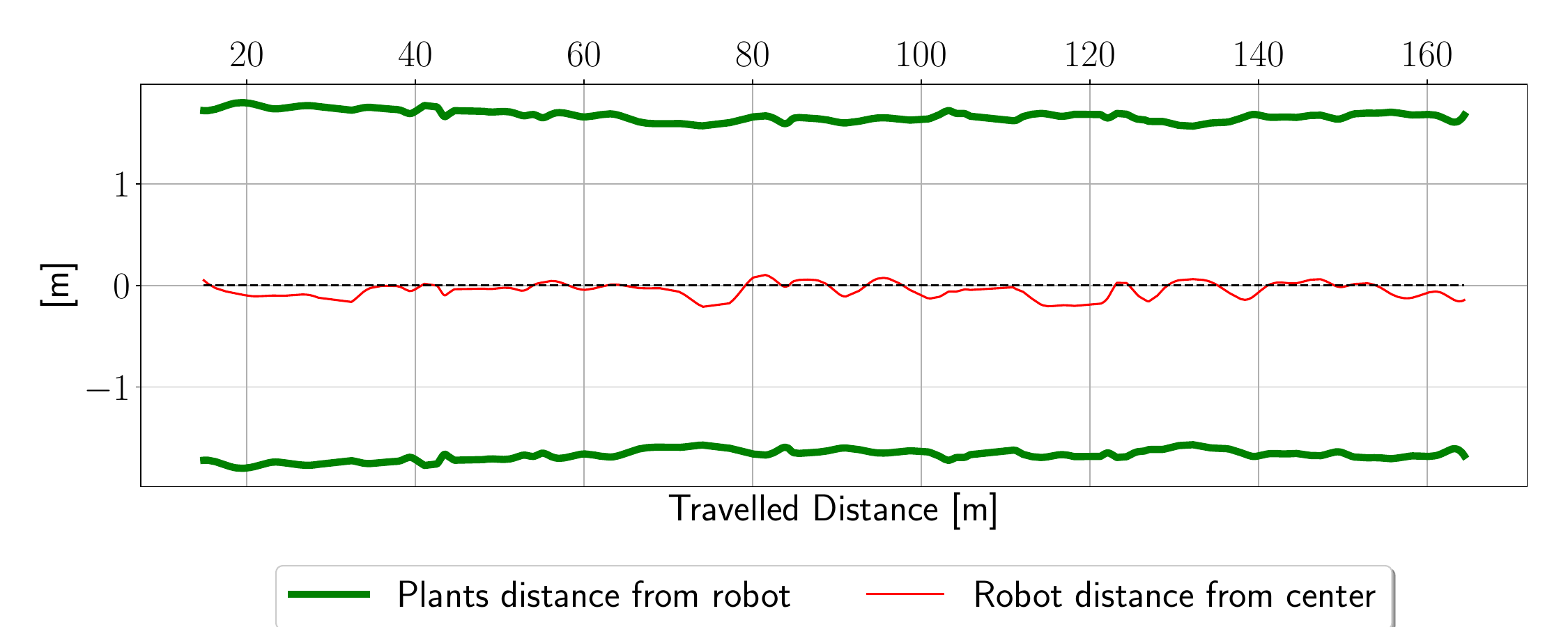}
    \includegraphics[width=.9\linewidth]{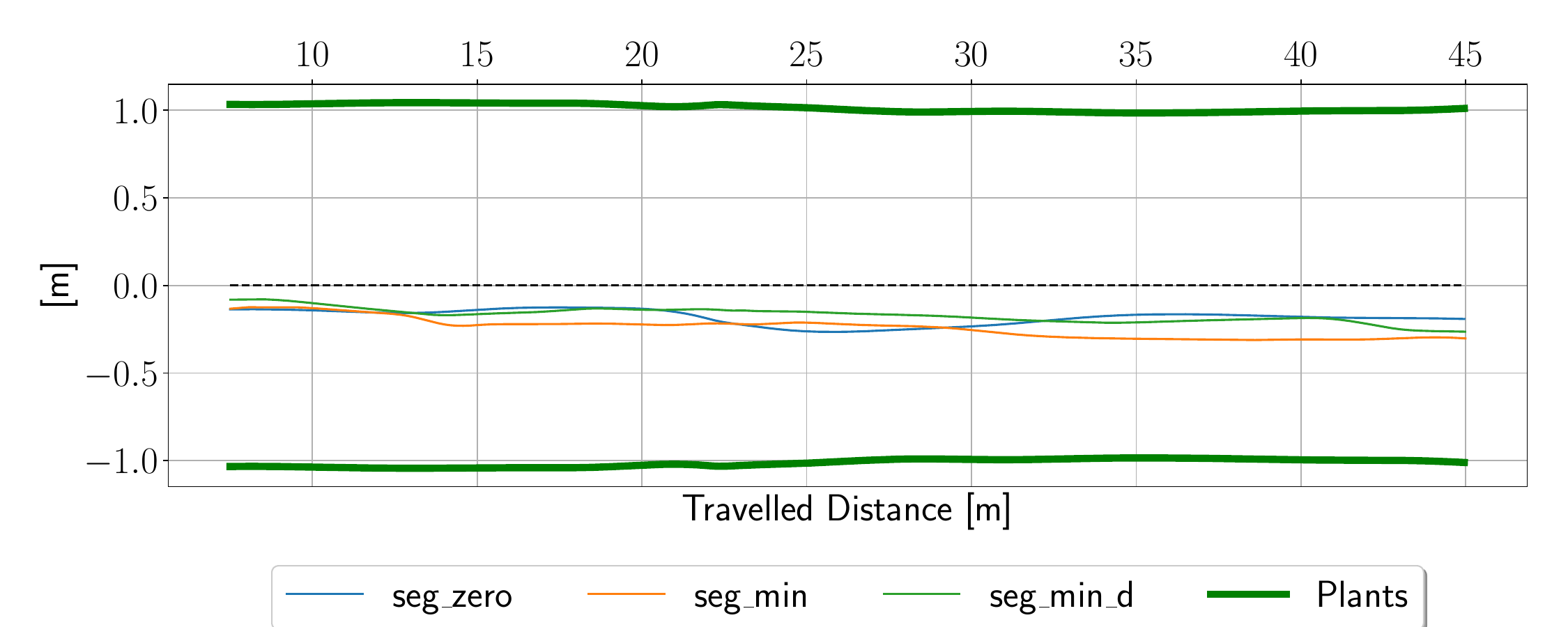}
    \caption{Trajectory results of relevant tests performed on the field. In order from top to bottom: navigation in pear tree rows using SegMinD, navigation in apple tree rows using SegMinD, and trajectory comparison of all three algorithms in a vineyard row. Sudden drifts in orchards traversal are caused by fruits, small obstacles, and irregularity in the terrain.}
    \label{fig:real-test-plot}
\end{figure}

The performances of the proposed control have been evaluated on two different plant orchards, i.e., apples and pears, and a vineyard. The performance metrics are reported in Table \ref{tab:algo-comp}. The trajectories of the best test for each crop field and the comparison between the proposed algorithms, SegMin, SegMinD, and SegZeros, are represented in Figure \ref{fig:real-test-plot}. Overall, the novel control laws can effectively solve the problem of guiding the robot through tree rows with thick canopies (high trees and pears) without a localization system in a real-world scenario. Moreover, the algorithms SegMin and SegMinD demonstrate the ability to generalize to the common case without obstruction by canopies. As shown by the comparison performed in the vineyard, they obtained results in line with the existing SegZeros. The algorithms SegMin and SegMinD show their effectiveness in maintaining the robot on the desired central line, even recovering from strong disturbances. As can be noticed by the trajectories in pear and apple trees (top and central plots in Figure \ref{fig:real-test-plot}), sudden drifts of the robot are caused by fruits, branches, stones, and disparate irregularity of the terrain. Those small obstacles cannot be precisely sensed and tackled with classic obstacle avoidance algorithms; hence, the resilience of the control algorithms to these external factors is crucial to keep the robot on track. Differently, in vineyard rows, grass and cleaner terrain induce smoother overall trajectories.

Finally, an ablation study is carried out on a vineyard row to assess the impact of key parameters on the proposed control strategy within the novel SegMinD algorithm. Specifically, the study explores the effects of the depth image max distance and pixel-wise confidence threshold of the predicted segmentation mask. The findings, detailed in Table \ref{tab:ablation-study}, indicate that a confidence level greater than 0.5 is required for achieving robust behavior, filtering mask portions with uncertain prediction. Indeed, results with a confidence level of 0.3 exhibit a high standard deviation in the angular velocity command. Regarding the depth image maximum distance, three values are tested: $5$, $8$, and $11$ m. A low value of $5$ m produces sub-optimal results compared to an intermediate value of $8$ m. In this scenario, the noise in the segmentation has a more pronounced effect as the long-view geometry of the row is not considered in the computation of the histogram, including only close plants. Conversely, a high value for the depth threshold leads to inferior results due to the insufficient precision of the depth camera, resulting in artifacts that can compromise overall performance. In conclusion, the optimal outcome is achieved with a high confidence in the prediction and an intermediate depth threshold of $8$ m.

\begin{table}[ht!]
    \centering
    \caption{Ablation study on SegMinD algorithm performance: relevant parameters of the segmentation control system are explored for a better understanding of their impact on the overall result. Three values of depth threshold and prediction confidence are selected for the ablation.}
    \label{tab:ablation-study}
    \begin{tabular}{@{}lcccc@{}}
    \toprule
    \textbf{Depth threshold} & \textbf{MAE} & \textbf{RMSE} & \textbf{Cum. $\pmb{\gamma_{avg}}$ [rad]} &  \textbf{$\omega$ STD [rad/s]} \\ \midrule
    \multicolumn{5}{c}{Confidence 0.3} \\ \midrule
    5.0 & 0.35234 & 0.36031 & -0.27982 & 0.99039 \\
    8.0 & 0.22854 & 0.23814 & -0.01200 & 0.25087 \\
    11.0 & 0.35295 & 0.36281 & 0.06704 & 0.38422 \\ \midrule
    \multicolumn{5}{c}{Confidence 0.5}  \\ \midrule
    5.0 & 0.22456 & 0.23915 & 0.02729 & 0.42077 \\
    8.0 & 0.15794 & 0.17106 & -0.19994 & 0.40949 \\
    11.0 & 0.45508 & 0.45754 & 0.00374 & 0.10303 \\ \midrule
    \multicolumn{5}{c}{Confidence 0.7}  \\ \midrule
    5.0 & 0.38653 & 0.39433 & 0.02949 & 0.46574 \\
    \textbf{8.0} & \textbf{0.11928} & \textbf{0.15057} & \textbf{-0.04034} & \textbf{0.12565} \\
    11.0 & 0.39656 & 0.40279 & -0.01696 & 0.35989 \\ \bottomrule
    \end{tabular}
\end{table}

\section{Conclusion}
\label{sec:conclusion} 
In this work, we presented a novel method to guide a service-autonomous platform through crop rows where a precise localization signal is often occluded by vegetation. Trees rows represented an open problem in row crop navigation since previous works based on image segmentation or processing failed due to the presence of branches and canopies covering the free passage for the rover in the image. The proposed pipeline SegMin and SegMinD overcome this limitation by introducing a global minimum search on the sum histogram over the mask columns. The experiments conducted demonstrate the ability to solve the navigation task in wide and narrow tree rows and, nonetheless, the improvement in efficiency and robustness provided by our method over previous works in generic vineyards scenarios. Moreover, real-world tests proved the reliable generalization properties of the efficient semantic segmentation neural network trained with synthetic data only. 

Future work will see the extension of the robot's capabilities to support agricultural tasks where more complex multi-objective behaviors are required, such as plant approach, box transport, and harvesting.

\subsection*{Acknowledgements} This work has been developed with the contribution of Politecnico di Torino Interdepartmental Center for Service Robotics PIC4SeR \footnote{\url{www.pic4ser.polito.it}}.

\bibliographystyle{unsrt}  
\bibliography{references}  


\end{document}